\documentclass[runningheads]{llncs}
\usepackage[T1]{fontenc}
\usepackage{graphicx,verbatim}
\usepackage{amsmath,amsfonts}
\usepackage{multirow} 
\usepackage{booktabs}
\usepackage{mathrsfs}
\usepackage{cite}
\usepackage[table]{xcolor}

\usepackage{booktabs}
\usepackage[normalem]{ulem}
\usepackage{makecell}
\usepackage{url} 
\begin{document}
\title{Counterfactual Contrastive Analysis}
%

\author{Yunlong He \and Pietro Gori\thanks{Corresponding author.}}


\authorrunning{Y. He and P. Gori}

\institute{
LTCI, Télécom Paris, Institut Polytechnique de Paris, France\\
\email{pietro.gori@telecom-paris.fr}
}

\maketitle              
\begin{abstract}
Visual Counterfactual Explanations (VCEs) aim to explain image classifiers by generating minimally edited and realistic versions of an input image that change the classifier’s prediction. Existing VCE methods are inherently classifier-dependent and therefore susceptible to classifier biases and failure modes, such as sensitivity to shortcut features and calibration errors.
In this paper, we propose a classifier-free approach for visual counterfactual generation based on Contrastive Analysis (CA). Given two datasets corresponding to different classes (e.g., healthy and patients), we disentangle the generative factors that are common across the two datasets from those that are salient to each dataset, and generate counterfactual images by swapping only the salient factors. By operating directly on data distributions rather than decision boundaries, our method provides model-agnostic VCEs that are less sensitive to classifier biases.
Our approach leverages the high-quality synthesis and well-structured latent space of StyleGAN2. We use the feature space $F$, instead than the usual $W$-space, to improve detail preservation. Unlike conventional CA approaches, which typically assume salient factors in only one dataset, we introduce an adapted framework and loss functions for VCE that allow multiple salient factors in each dataset. We evaluate our method on three medical imaging datasets and demonstrate superior counterfactual generation quality compared to existing approaches.
\keywords{Visual Counterfactual Explanations  \and Contrastive Analysis}
%
\end{abstract}

\section{Introduction}
With the rapid scaling of modern AI models, the ability to explain black-box decisions has become critical, particularly in medical imaging. Explainability often governs whether models can be trusted and accepted for real-world use. Visual counterfactual explanations (VCEs) \cite{jeanneret2022diffusion,augustin2022diffusion,goyal_counterfactual_2019,Jeanneret_2023_CVPR} are widely studied as a way to probe a classifier’s behavior, seeking the smallest realistic changes to an input image that would alter its prediction.

VCEs for image classifiers are typically realized by generating a counterfactual (CF) image as an edited version of the original input. This task is often posed as a constrained generation problem with three objectives. First, the CF must be \textbf{valid}, meaning that it changes the classifier’s prediction to the desired outcome. Second, the edited image should remain \textbf{realistic}, staying on the image manifold and avoiding unnecessary changes to irrelevant content. Third, the induced changes should be \textbf{interpretable} and \textbf{minimal} in a \emph{semantic} sense, so that the explanation clearly indicates the key evidence responsible for the classifier's decision change. 
Most VCE methods instantiate these objectives within a generative framework, leveraging VAEs~\cite{rodriguez2021beyond}, GANs~\cite{lang2021explaining,singla2019explanation}, or diffusion models (DMs)~\cite{Jeanneret_2023_CVPR,jeanneret2022diffusion,weng2024fast,jeanneret2024text}. Among them, DMs currently achieve state-of-the-art performance for realistic CF image generation, typically guiding the generation process with classifier-driven losses to reach a label-flipping outcome, while penalizing deviations to preserve plausibility and similarity to the input. However, classifier guidance can induce changes aligned with the classifier biases and failure modes, such as sensitivity to shortcut features \cite{weng2024fast} and calibration errors, rather than realistic changes in the underlying data distribution, potentially reducing edit fidelity. In addition, these methods offer limited insight into which semantic factors drive the prediction flip, thereby limiting their interpretability.

In this paper, we introduce a classifier-free approach for visual counterfactual generation grounded in the \emph{Contrastive Analysis} (CA) framework~\cite{abid2018exploring,weinberger2022moment,carton2024double,louiset24a,louiset2024sepclr,he2025learningcommonsalientgenerative}. Instead of generating CFs based on a classifier’s decision boundary, we operate at the data-distribution level, as in CA. Given two class-specific datasets, the proposed method aims to uncover the common and salient (i.e., class-specific) generative factors. CFs can then be generated by swapping only the salient factors, while preserving the common ones.  
Most CA methods are based on deep generative models, such as VAEs~\cite{bousmalis2016domain,weinberger2022moment,severson2019unsupervised,zou2022joint,benaim2019domain,kleinman2023gacskorner} and GAN\cite{carton2024double, gonzalez2018image}. While reconstruction can regularize the salient space by suppressing common information, it is insufficient on its own and typically requires additional constraints for clean latent separation \cite{weinberger2022moment, louiset24a, louiset2024sepclr, sanchez2020disentangled}.
A recent work \cite{he2025learningcommonsalientgenerative} extended CA to high-quality image datasets using modern GANs and diffusion models and reported that StyleGAN2 \cite{Karras2019stylegan2} achieves diffusion-level quality while enabling substantially faster inference. In parallel, advances in StyleGAN editing have shown that translating edits from the W-space to the generator’s intermediate feature space (F-space) improves fine-grained details \cite{bobkov2024devil}. 

Inspired by these findings, we propose a StyleGAN-based CA framework that leverages latent disentanglement and F-space refinement for VCEs on medical imaging datasets. Our contributions are:
\begin{enumerate}
\item A classifier-free VCE generative framework grounded in CA that enables controllable, classifier-agnostic CF generation via disentangled latent representations, yielding high-fidelity results and high semantic interpretability.
\item A CA-based framework, supporting both background-target and multi-salient assumptions, based on StyleGAN2’s high-quality synthesis and well-structured F-space, instead of standard W-space, for better detail preservation.
\item Experiments on three medical imaging datasets demonstrate improved disentanglement performance over CA baselines and superior generation quality to prior state-of-the-art VCE approaches with faster image editing (0.25s/image).
\end{enumerate}

\begin{figure}[t]
    \centering
    \includegraphics[width=0.9\linewidth]{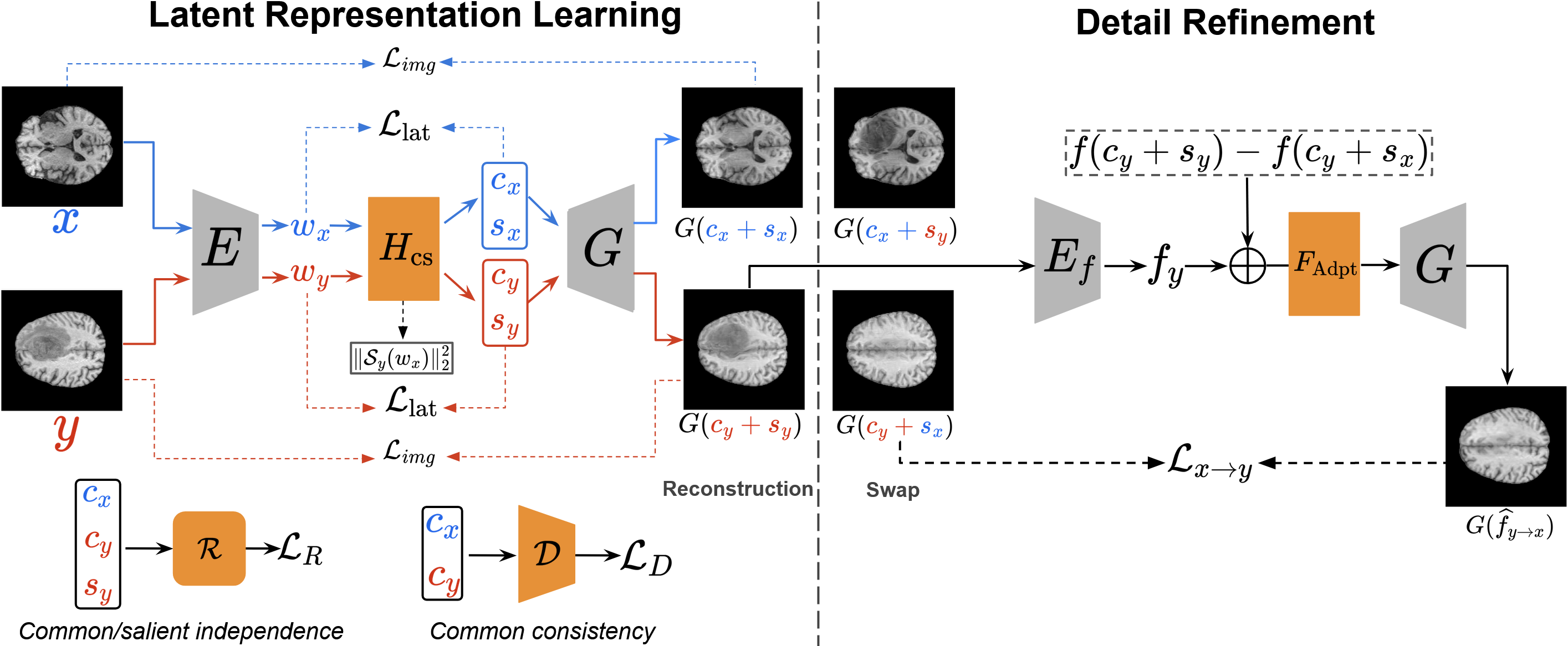}
    \caption{\textbf{Overview of the proposed CA-based framework for CF generations.} \underline{Left}: we first disentangle each input into a common $c$ and a salient factor $s$, enabling reconstruction and CF editing by swapping the salient factors between images from $X$ and $Y$. \underline{Right}: we then refine the synthesized image by adjusting $F$-space features according to the feature shift $\Delta f_{y \rightarrow x} = f(c_y + s_y) - f(c_y + s_x)$ common-salient factor manipulation.}
    \label{fig:architecture}
\end{figure}

\section{Method}
\label{sec:methods}

\noindent \textbf{Problem Statement.} In the CA setting, we consider two datasets $X=\{x_i\}_{i=1}^{N}$ and $Y=\{y_j\}_{j=1}^{M}$. 
Most prior work adopts a \textit{background-target} (BT) assumption, where only one dataset (typically $Y$) contains a dataset-specific (or \textit{salient}) pattern (e.g., a pathology) absent from the other, while both share common content (e.g., healthy anatomy). We generalize CA to a \textit{multiple-salient} (MS) setting, where both $X$ and $Y$ contain their own \textit{salient} patterns that are absent in the other. 
Our goal is to learn a \textit{common} generative factor $c \in \mathbb{R}^{d_c}$ capturing information shared across $X$ and $Y$, and \textit{salient} generative factors $s_x \in \mathbb{R}^{d_{sx}}$, $s_y \in \mathbb{R}^{d_{sy}}$ that capture variations unique to $X$ and $Y$, respectively. 


\noindent \textbf{Common and Salient (CS)-StyleGAN.} In the context of StyleGAN, given samples $x \in X$ and $y \in Y$, an encoder $E$ (i.e., pSp~\cite{richardson2021encoding}) maps them to latent codes $w_x=E(x)$ and $w_y=E(y)$, where $w_x,w_y \in \mathbb{R}^{d_w}$.
We introduce a CS separator $H_{\mathrm{cs},\phi}$, which consists of one common branch $\mathcal{C}_{\phi_c}$ and two salient branches $\mathcal{S}_{x,\phi_{sx}}$ and $\mathcal{S}_{y,\phi_{sy}}$. For brevity, we omit the parameters $\phi$ in the following. We define the latent factors as: $c_x=\mathcal{C}(w_x)$ and $c_y=\mathcal{C}(w_y)$. In the MS setting, we set $s_x=\mathcal{S}_{x}(w_x)$ and $s_y=\mathcal{S}_{y}(w_y)$, while in the BT setting, we remove $\mathcal{S}_{x}$ and use $\mathcal{S}_{y}$ only (all $\mathcal{S}_{x}$-related terms are omitted).

\noindent \textbf{Latent Regularization.}
We first train the $H_{cs}$ separator with the following reconstruction loss in the StyleGAN2 latent space:
\begin{equation}
\mathcal{L}_{\text{lat}} =
\| \mathcal{C}(w_x) + \mathcal{S}_{x}(w_x) - w_x \|_2^2
+
\| \mathcal{C}(w_y) + \mathcal{S}_{y}(w_y) - w_y \|_2^2,
\label{equ:loss_lat}
\end{equation}
where each term encourages the sum of the common output and the corresponding salient output to reconstruct the StyleGAN2 latent ($w_x$ or $w_y$). To enforce that the salient factors capture only dataset-specific information, we additionally penalize cross-dataset activations, by minimizing $\|\mathcal{S}_{y}(w_x)\|_2^2$ and $\|\mathcal{S}_{x}(w_y)\|_2^2$ (we use only $\|\mathcal{S}_{y}(w_x)\|_2^2$ under the BT assumption, as explained in Fig.~\ref{fig:architecture}). 

The loss in Eq.~\eqref{equ:loss_lat} is necessary but insufficient. In particular, it does not enforce (i) \emph{common consistency}, i.e., that the common factors capture the same information across $X$ and $Y$, nor (ii) \emph{common-salient independence}, i.e., that common and salient factors do not share information. To promote \emph{common consistency}, we impose distributional alignment of the common factors by introducing a discriminator $\mathcal{D}$ that predicts whether $\mathcal{C}(w)$ originates from $X$ or $Y$, while the separator $H_{\mathrm{cs}}$ is trained to make this prediction impossible:
\begin{equation}
\mathcal{L}_{D}
= \min_{H_{\mathrm{cs}}} \max_{\mathcal{D}}
\Big(
- \mathbb{E}_{w_y}[\log \mathcal{D}(\mathcal{C}(w_y))]
- \mathbb{E}_{w_x}[\log (1-\mathcal{D}(\mathcal{C}(w_x)))]
\Big).
\label{equ:loss_L_D}
\end{equation}
Here, $\mathcal{D}$ is optimized to discriminate $X$ from $Y$ using the common codes, whereas $H_{\mathrm{cs}}$ is optimized adversarially so that $\mathcal{C}(w_x)$ and $\mathcal{C}(w_y)$ become indistinguishable. 

To encourage \textbf{\emph{common--salient statistical independence}}, we introduce two regressors, $\mathcal{R}_{x}$ and $\mathcal{R}_{y}$, that attempt to predict the salient outputs $\mathcal{S}_{x}(w_x)$ and $\mathcal{S}_{y}(w_y)$ from the corresponding common outputs $\mathcal{C}(w_x)$ and $\mathcal{C}(w_y)$, respectively. 
If the regressors can accurately predict salient output from the common one, this indicates that $\mathcal{C}(w)$ still contains salient information and the separation is incomplete. We therefore train the separator adversarially against the regressors by minimizing the following objective:
\begin{equation}
\begin{aligned}
\mathcal{L}_{R}
=
\min_{H_{\mathrm{cs}}}\max_{\mathcal{R}}
\Big(\!
&-\mathbb{E}_{w_x}\!\big[\|\mathcal{R}_{x}(\mathcal{C}(w_x))\!-\!\mathcal{S}_{x}(w_x)\|_2^2\big]\!-\mathbb{E}_{w_y}\!\big[\|\mathcal{R}_{x}(\mathcal{C}(w_y))\big] \\
&-\mathbb{E}_{w_y}\!\big[\|\mathcal{R}_{y}(\mathcal{C}(w_y))\!-\!\mathcal{S}_{y}(w_y)\|_2^2\big]\!-\mathbb{E}_{w_x}\!\big[\|\mathcal{R}_{y}(\mathcal{C}(w_x))\big]
\Big).
\end{aligned}
\label{equ:loss_L_R}
\end{equation}
This  discourages $\mathcal{C}(w)$ from encoding salient patterns. We regress $s$ from $c$ because $c$  is typically bigger (more information) 
and thus tends to absorb $s$. 

\noindent \textbf{Image-Space Losses.}
We adopt an image-space reconstruction loss combining pixel and perceptual terms. Given a real image $i$ and its reconstruction $\widehat{i}$, we use $\mathcal{L}_{\mathrm{rec}}(i, \hat{i}) = \|i - \hat{i}\|_2^2 + \|V(i) - V(\hat{i})\|_2^2$,
where the first term is a pixel-wise $\ell_2$ loss, and the second term is the LPIPS perceptual loss~\cite{zhang2018unreasonable}. The function $V(\cdot)$ extracts perceptual features using a pretrained VGG16 network~\cite{simonyan2015vgg}. We use this Eq. to reconstruct both $x$ and $y$, minimizing the image-space loss:
\begin{equation}
    \mathcal{L}_{\text{img}}= \mathcal{L}_{\mathrm{rec}}(x, G(c_{x} + s_{x})) + \mathcal{L}_{\mathrm{rec}}(y, G(c_{y} + s_{y})),
    \label{equ:loss_img}
\end{equation}
where $G$ is the pretrained StyleGAN2 generator.
The final loss to train $H_{cs}$ is:
\begin{equation}
    \mathcal{L}_{H_{cs}} =\lambda_{lat} \mathcal{L}_{\mathrm{lat}} + \lambda_{D}\mathcal{L}_{D} +  \lambda_{R}\mathcal{L}_{R} +   \lambda_{img}\mathcal{L}_{\text{img}}
    \label{equ:loss_separator}
\end{equation}

\noindent \textbf{F-space Refinement.} 
We first learn common and salient factors in the $W$ space of StyleGAN2. Then, we add a refinement module that edits intermediate StyleGAN feature space (F space), as in \cite{bobkov2024devil}.
Let $f(\cdot)$ denote an intermediate feature of the generator $G$, e.g., the output of layer $l{=}9$.
Given the learned latent factors, we define F-space feature shifts for the two swap directions as:
\begin{equation}
\Delta f_{x \rightarrow y} = f(c_x + s_x) - f(c_x + s_y); \;\;
\Delta f_{y \rightarrow x} = f(c_y + s_y) - f(c_y + s_x).
\label{equ:delta_f}
\end{equation}
Intuitively, $\Delta f_{x \rightarrow y}$ captures the change induced by replacing the salient factor of $x$ with the one of $y$ while keeping $c_x$ fixed, and $\Delta f_{y \rightarrow x}$ is defined analogously.

\noindent To condition the shifts on the source image, we extract features from reconstructions using a pretrained F-space encoder $E_f$: $f_x\! =\! E_f(G(c_x\!+\! s_x))$ and $f_y \!=\! E_f(G(c_y \!+\! s_y))$.
As in~\cite{bobkov2024devil}, we concatenate $f$ and $\Delta f$, and pass them to an adapter $F_{\text{Adpt}}$, yielding: $\widehat{f}_{x \! \rightarrow \!y} \!= \!F_{\text{Adpt}}(f_x \!\oplus \!\Delta f_{x \!\rightarrow \!y})$ and $\widehat{f}_{y\! \rightarrow \!x} \!= \!F_{\text{Adpt}}(f_y \!\oplus \!\Delta f_{y \!\rightarrow \!x})$,
where $\oplus$ denotes concatenation. Decoding the adapted features with $G$ yields the edited images. $F_{\text{Adpt}}$ is trained with:
$\mathcal{L}_{x \rightarrow y} = \mathcal{L}_{\mathrm{rec}}(G(c_x + s_y), \; G(\widehat{f}_{x \rightarrow y}))$
and
$\mathcal{L}_{y \rightarrow x} = \mathcal{L}_{\mathrm{rec}}(G(c_y + s_x), \; G(\widehat{f}_{y \rightarrow x}))$.
We also include a real-image reconstruction term to preserve performance on real inputs:
$\mathcal{L}_{\text{real}} = \mathcal{L}_{\mathrm{rec}}(x,\; \widehat{x}) + \mathcal{L}_{\mathrm{rec}}(y,\; \widehat{y})$,
where $\widehat{x}=G(F_\text{Adpt}(E_f(x)))$ and $\widehat{y}=G(F_\text{Adpt}(E_f(y)))$.
Finally, we encourage realism with an adversarial loss $\mathcal{L}_{\text{adv}}=\mathcal{L}_{\text{adv-}x}+\mathcal{L}_{\text{adv-}y}$, where $\mathcal{L}_{\text{adv-}x}$ is applied to $G(\widehat{f}_{y \rightarrow x})$ against $x$, and $\mathcal{L}_{\text{adv-}y}$ to $G(\widehat{f}_{x \rightarrow y})$ against $y$. 
The overall refinement objective is:
$\mathcal{L}_{\text{refine}} = \mathcal{L}_{x \rightarrow y} + \mathcal{L}_{y \rightarrow x} + \mathcal{L}_{\text{real}} + \lambda_{adv} \mathcal{L}_{adv}$.

\section{Experiments}
\label{sec:dataset}
\noindent \textbf{Datasets.} We evaluate our method on three 2D medical imaging datasets: \textbf{BloodMNIST}\cite{medmnistv2} (Eosinophil Vs Neutrophil; Train=2500/Test=500), \textbf{OCTMNIST}\cite{medmnistv2} (Normal Vs Choroidal neovascularization; Train=17000/Test=3000 + Diabetic macular edema Vs Drusen; Train=7000/Test=2000), and \textbf{BraTS2023} \cite{menze2015brats} (Healthy Vs Tumor; Train=8000/Test=2000). For each dataset, we define two groups, denoted $X$ and $Y$, based on the provided labels. We construct balanced splits such that $X$ and $Y$ contain the same number of images in both training and test sets. All images are preprocessed to a resolution of $256\times256$. For BloodMNIST and OCTMNIST, we resize the $224\times224$ images from the MedMNIST+ release to $256\times256$. For MRI scans, we use zero-padding to obtain $256\times256$ images while preserving the original aspect ratio.

\noindent \textbf{Implementation Details.}
We train the proposed CS-StyleGAN in two stages. In \textbf{Stage~1}, we optimize the separator $H_{\mathrm{cs}}$ to disentangle common and salient factors in the $\mathcal{W}^{+}$ space. We first warm up $H_{\mathrm{cs}}$ for 2,000 steps using only $\mathcal{L}_{\mathrm{lat}}$ and $\mathcal{L}_{\mathrm{img}}$ to obtain informative latent factors. We then alternate between updating $\mathcal{D}$ and $\mathcal{R}$ and updating $H_{\mathrm{cs}}$: with $H_{\mathrm{cs}}$ fixed, $\mathcal{D}$ and $\mathcal{R}$ are updated by maximizing the objectives in Eqs.~\eqref{equ:loss_L_D}--\eqref{equ:loss_L_R} using separate Adam optimizers with a learning rate of $10^{-4}$; with $\mathcal{D}$ and $\mathcal{R}$ fixed, $H_{\mathrm{cs}}$ is updated by minimizing Eq.~\eqref{equ:loss_separator} using an Adam optimizer with a learning rate of $10^{-3}$. In \textbf{Stage~2}, using the common and salient factors learned in Stage~1 and their corresponding F-space feature shifts, we optimize the refinement module ($F_{\mathrm{Adpt}}$) with $\mathcal{L}_{\mathrm{refine}}$. We use the Ranger optimizer employed in SFE~\cite{bobkov2024devil}, with our learning rate set to $10^{-4}$. For each dataset, the pSp encoder $E$, StyleGAN2 generator $G$, and F-space encoder $E_f$ are pretrained on the corresponding training set before the two-stage training procedure. Detailed network architectures are provided in the supplementary material. The implementation code is available at \url{https://github.com/BioMedTP/CF_Contrastive_Analysis}.

\noindent \textbf{Baselines and Evaluation Metrics.}
We compare our method with CA baselines MM-cVAE~\cite{weinberger2022moment}, SepVAE~\cite{louiset2024sepclr}, and Double InfoGAN~\cite{carton2024double}, and with diffusion-based VCE methods: ACE~\cite{Jeanneret_2023_CVPR}, DiME~\cite{jeanneret2022diffusion}, FastDiME~\cite{weng2024fast}, and TIME~\cite{jeanneret2024text}. We also consider off-the-shelf T2I DMs as CF alternatives (not CF-specific methods), including FLUX~\cite{blackforestlabs2025flux1kontext} and SDXL~\cite{podell2024sdxl} adapted with LoRA~\cite{hu2022lora}. We evaluate the proposed method on CA and CF generation using widely adopted metrics to evaluate synthesis quality: L2, LPIPS~\cite{zhang2018unreasonable}, MS-SSIM~\cite{wang2003msssim}, and FID~\cite{heusel2017gans}. Furthermore, we assess whether swapped outputs match the post-edit domain distribution using FID, reporting FID$_{X\!\to\!Y}$ (FID between $X$ images after swapping to $Y$ and real $Y$) and FID$_{Y\!\to\!X}$ (swapped $Y$ vs.\ real $X$). 
We evaluate editing success using the same U-Net-based classifier
architecture as in~\cite{weng2024fast}, trained on real images from the
corresponding training set. We report the classification accuracy with
respect to the desired post-edit label. About latent separation quality, we train a classifier on the learned latent factors to predict the class ($X$ or $Y$). Higher accuracy indicates the class is encoded in the corresponding (common or salient) space. Please note that in the BT assumption, the information about $X$ is entirely encoded in $c$. Eventually, following VCE prior works ~\cite{weng2024fast, Jeanneret_2023_CVPR}, we report FID, sFID, Flip Ratio (FR), Mean Absolute Difference (MAD) and Bidirectional KL Divergence (BKL) for measuring counterfactual quality.

\begin{figure}[t]
    \centering
\includegraphics[width=0.9\linewidth]{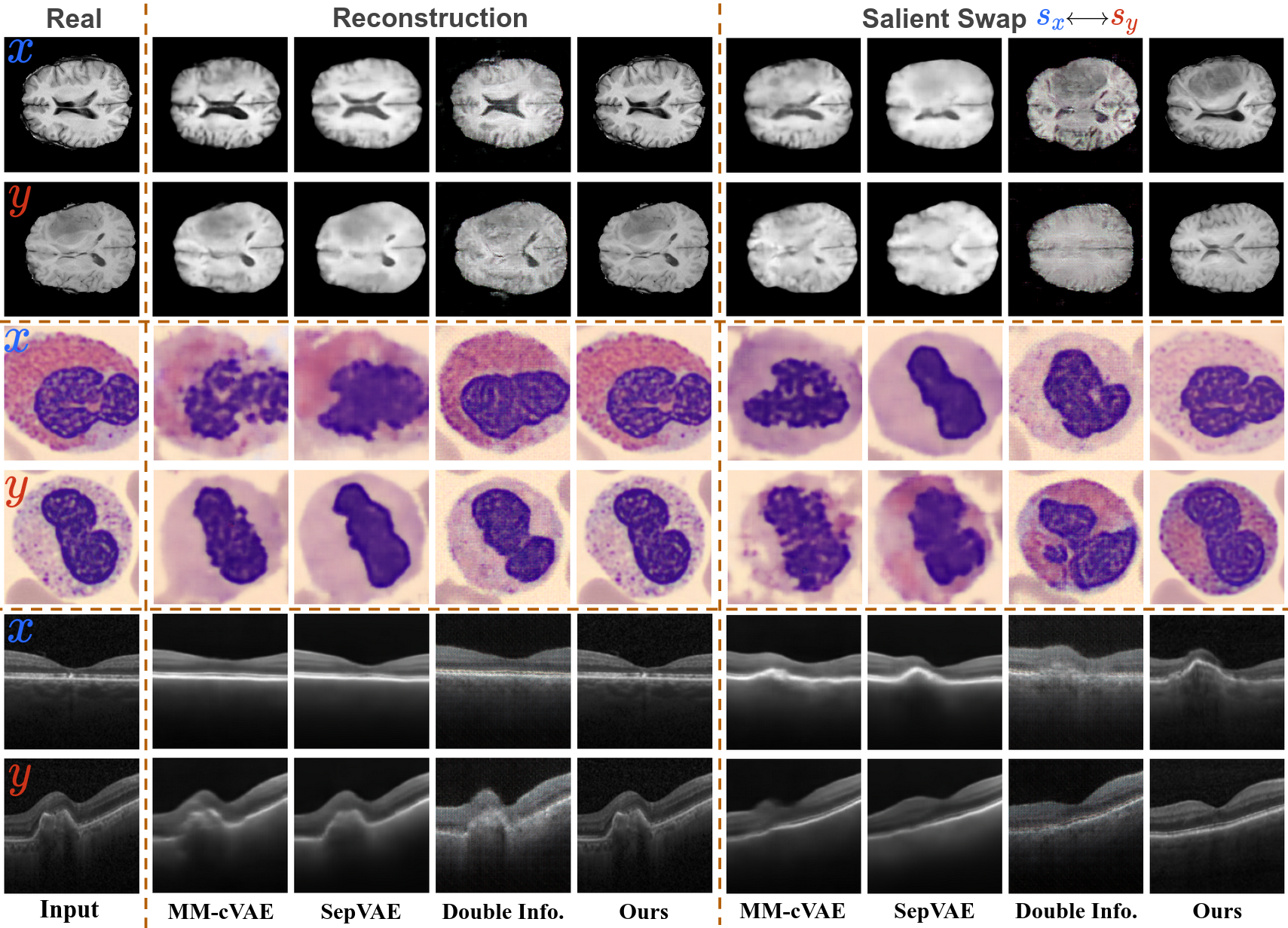}
    \caption{\textbf{Comparisons with CA baselines} on BraTS, BloodMNIST, and OCTMNIST (Normal vs.\ CNV) (top to bottom). 
    \emph{Reconstruction}: use the common and salient factors from original images. \emph{Salient Swap}: swap salient factors between $x$ and $y$ while keeping common factors fixed during generation.}
    \label{fig:baseline_compare}
\end{figure}

\begin{table}[t]
\caption{Quantitative comparison results on multiple datasets. Best results in bold, second-best underlined (within each dataset block). Ref: $F$-space refinement; BT/MS Ass.: background-target/multi-salient assumptions.}
\label{tab:ca_quality}
\centering
\footnotesize
\resizebox{0.85\textwidth}{!}{
\begin{tabular}{c|c|ccccc|ccccc|ccc}
\toprule
& & \multicolumn{5}{c|}{\textbf{Reconstruction ($X$)}} &
    \multicolumn{5}{c|}{\textbf{Reconstruction ($Y$)}} &
    \multicolumn{3}{c}{\textbf{Swap}} \\
\cmidrule(lr){3-7} \cmidrule(lr){8-12} \cmidrule(lr){13-15}
\textbf{Dataset ($X$ vs. $Y$)} & \textbf{Method}
& PSNR$\uparrow$ & MSE$\downarrow$ & MS-SSIM$\uparrow$ & LPIPS$\downarrow$ & FID$\downarrow$
& PSNR$\uparrow$ & MSE$\downarrow$ & MS-SSIM$\uparrow$ & LPIPS$\downarrow$ & FID$\downarrow$
& FID$_{X\!\to\!Y}\downarrow$ & FID$_{Y\!\to\!X}\downarrow$
& Acc (X\!+\!Y)$\uparrow$ \\
\midrule

\multirow{5}{*}{\makecell{\textbf{BraTS}\\\textbf{(Healthy vs. Tumored)}}}
& \textbf{MMc-VAE}        & 20.32 & 0.008 & 0.852 & 0.163 & 89.86 & 21.06 & 0.007 & 0.951 & 0.157 & 86.36 &  98.63 & 92.77 & 0.753 \\
& \textbf{SepVAE}         & 23.24 & 0.006 & 0.933 & 0.156 & 115.2 & 23.75 & 0.006 & 0.930 & 0.152 & 109.4 &  113.7 & 111.9 & 0.821 \\
& \textbf{Double InfoGAN} & 23.96 & 0.005 & 0.925 & 0.151 & 125.1 & 25.44 & 0.004 & 0.958 & 0.145 & 128.3 &  137.2 & 135.4 & 0.901 \\
& \textbf{Ours (pSp-cs)}  & \uline{24.03} & \uline{0.004} & \uline{0.91} & \uline{0.054} & \uline{44.46}
                          & \uline{24.75} & \uline{0.003} & \uline{0.90} & \uline{0.055} & \uline{30.10}
                          & \uline{37.10} & \uline{45.50} & \textbf{0.939} \\
& \textbf{Ours (pSp-cs-Ref)} & \textbf{38.19} & \textbf{0.0002} & \textbf{0.99} & \textbf{0.004} & \textbf{2.64}
                             & \textbf{38.20} & \textbf{0.0002} & \textbf{0.99} & \textbf{0.005} & \textbf{2.57}
                             & \textbf{25.17} & \textbf{32.56} & \uline{0.914} \\
\midrule


\multirow{5}{*}{\makecell{\textbf{BloodMNIST}\\\textbf{(Eos. vs. Neutrophil)}}}
& \textbf{MMc-VAE}        & 18.27 & 0.015 & 0.556 & 0.378 & 175.1 & 19.42 & 0.011 & 0.643 & 0.357 & 179.6 & 181.7 & 183.9 & 0.819 \\
& \textbf{SepVAE}         & 18.04 & 0.016 & 0.551 & 0.421 & 182.8 & 18.01 & 0.016 & 0.579 & 0.419 & 219.5 & 217.5 & 221.9 & 0.691 \\
& \textbf{Double InfoGAN} & 19.45 & 0.015 & 0.567 & 0.472 & 95.97 & 19.70 & 0.014 & 0.606 & 0.491 & 104.7 & 149.50 & 135.7 & 0.646 \\
& \textbf{Ours (pSp-cs)}  & \uline{22.35} & \uline{0.006} & \uline{0.725} & \uline{0.139} & \uline{73.46}
                          & \uline{22.43} & \uline{0.006} & \uline{0.760} & \uline{0.139} & \uline{70.42}
                          & \uline{84.05} & \uline{88.61} & \uline{0.979} \\
& \textbf{Ours (pSp-cs-Ref)} & \textbf{41.99} & \textbf{0.0001} & \textbf{0.998} & \textbf{0.004} & \textbf{2.887}
                             & \textbf{42.54} & \textbf{0.0001} & \textbf{0.998} & \textbf{0.004} & \textbf{2.590}
                             & \textbf{43.37} & \textbf{45.90} & \textbf{0.981} \\
\midrule

\multirow{7}{*}{\makecell{\textbf{OCTMNIST}\\\textbf{(DME vs. Drusen)}}}
& \textbf{MMc-VAE}        & 23.44 & 0.005 & 0.805 & 0.404 & 128.05 
                          & 26.19 & 0.003 & 0.867 & 0.390 & 117.71 
                          & 126.49 & 128.17 & 0.665 \\

& \textbf{SepVAE}         & 23.89 & 0.005 & 0.822 & 0.400 & 107.84 
                          & 26.26 & 0.003 & 0.872 & 0.395 & 105.92 
                          & 131.22 & 143.67 & 0.597 \\

& \textbf{Double InfoGAN} & 16.84 & 0.026 & 0.592 & 0.369 & 181.38 
                          & 18.89 & 0.017 & 0.698 & 0.300 & 179.87 
                          & 189.06 & 199.42 & 0.832 \\

& \textbf{Ours (BT Ass.)}      & 23.79 & 0.005 & 0.806 & 0.111 & 51.10 
                          & 24.99 & 0.004 & 0.830 & 0.084 & 42.62 
                          & 65.66 & 79.13 & 0.627 \\

& \textbf{Ours (BT Ass. Ref)}  & \textbf{38.24} & \uline{0.0002} & \uline{0.992} & \uline{0.0143} & \uline{5.832}
                          & \uline{39.02} & \uline{0.0001} & \uline{0.994} & \uline{0.012} & \textbf{3.294}
                          & \uline{44.67} & \uline{63.62} & 0.689 \\

& \textbf{Ours (MS Ass.)}   & 26.50 & 0.005 & 0.814 & 0.108 & 48.64
                          & 24.70 & 0.004 & 0.837 & 0.083 & 39.43
                          & 59.21 & 73.79 & \textbf{0.876} \\

& \textbf{Ours (MS Ass. Ref)} 
                          & \uline{38.09} & \textbf{0.0001} & \textbf{0.994} & \textbf{0.014} & \textbf{5.356}
                          & \textbf{39.00} & \textbf{0.0001} & \textbf{0.995} & \textbf{0.012} & \uline{3.911}
                          & \textbf{39.24} & \textbf{55.92} & \uline{0.863} \\

\bottomrule
\end{tabular}
}
\end{table}

\noindent \textbf{Comparison with CA Baselines.}
Fig.~\ref{fig:baseline_compare} qualitatively compares our method with CA baselines for reconstruction and salient swapping. Our reconstructions are sharper and better preserve fine anatomical structures.
For salient swapping, our approach transfers the salient pattern (tumors in BraTS, eosinophil–neutrophil staining differences in BloodMNIST, and CNV-related abnormalities in OCTMNIST) while preserving shared anatomy (healthy brain tissues, global cell layout/nuclear morphology, and normal retinal layer structure, respectively). Table~\ref{tab:ca_quality} reports quantitative results, where our pSp-based method, with or without F-space refinement (\textbf{-Ref}), consistently outperforms all CA baselines on both reconstruction and swapping. Notably, with the MS setting, our method better handles the challenging DME vs.\ Drusen case, achieving the best reconstruction metrics and the highest swapping quality (FID and classification accuracy of 0.876). Finally, Table~\ref{tab:ca_sep} evaluates latent separation between the learned common ($C$) and salient ($S$) factors. Salient patterns are mainly captured in $S$ rather than in $C$, and our method attains the lowest separation gap $\Delta$ across all methods, indicating improved semantic disentanglement over prior approaches. In Table \ref{tab:ablation}, we also report an ablation study.

\begin{table}[t]
\centering
\begin{minipage}{0.50\textwidth}
\centering
\caption{Latent separation results. Mean $\pm$ std of 5-fold cross-validated classification accuracy to distinguish between the class of $X$ or $Y$ using logistic regression and using as features either $C$, $S_x$ or $S_y$. 
$\Delta = \lvert 0.5 - C \rvert + \lvert 1.0 - S_1 \rvert (+ \lvert 1.0 - S_2 \rvert$).}
\label{tab:ca_sep}
\resizebox{\textwidth}{!}{%
\begin{tabular}{l|*{3}{c}|*{4}{c}}
\toprule
& \multicolumn{3}{c|}{\textbf{BraTS (Healthy vs. Tumored)}} &
  \multicolumn{4}{c}{\textbf{OCTMNIST (DME vs. Drusen)}} \\
\cmidrule(lr){2-4}\cmidrule(lr){5-8}
\textbf{Model} 
& $C$ & $S_y$ & $\Delta$
& $C$ & $S_x$ & $S_y$ & $\Delta$ \\
\midrule

\textbf{MM-cVAE}
  & 0.68 $\pm$ 0.02 & 0.73 $\pm$ 0.02 & 0.45
  & 0.61 $\pm$ 0.01 & 0.69$\pm$ 0.03 & -- & 0.42 \\

\textbf{SepVAE} 
  & 0.67 $\pm$ 0.02 & 0.92 $\pm$ 0.01 & 0.25
  & 0.64 $\pm$ 0.02 & 0.94 $\pm$ 0.01 & -- & 0.20 \\

\textbf{Double Info.} 
  & 0.65 $\pm$ 0.01 & 0.86 $\pm$ 0.01 & 0.29
  & 0.76 $\pm$ 0.02 & 0.61 $\pm$ 0.02 & -- & 0.65 \\

\textbf{Ours (BT Ass.)} 
  & \textbf{0.58 $\pm$ 0.08} & \textbf{0.95 $\pm$ 0.01} & \textbf{0.13}
  & 0.57 $\pm$ 0.08 & 0.96 $\pm$ 0.006 & -- &  0.11 \\

\textbf{Ours (MS Ass.)} 
  & -- & -- & --
  & \textbf{0.54 $\pm$ 0.05} & \textbf{0.97 $\pm$ 0.008} & \textbf{0.98} $\pm$ \textbf{0.003} & \textbf{0.09} \\

\textbf{\textcolor{blue}{Expected}} &
\textcolor{blue}{0.5} & \textcolor{blue}{1.0} & \textcolor{blue}{0} &
\textcolor{blue}{0.5} & \textcolor{blue}{1.0} & \textcolor{blue}{1.0} & \textcolor{blue}{0} \\
\bottomrule
\end{tabular}%
}
\end{minipage}
\hfill
\begin{minipage}{0.45\textwidth}
\centering
\caption{Ablation study on the BraTS dataset: Healthy (X) vs. Tumored (Y).}
\label{tab:ablation}
\resizebox{\textwidth}{!}{%
\begin{tabular}{l|*{3}{c}|*{3}{c}}
\toprule
& \multicolumn{3}{c|}{\textbf{Latent Separation}} &
  \multicolumn{3}{c}{\textbf{Image Edit Quality}} \\
\cmidrule(lr){2-4}\cmidrule(lr){5-7}
\textbf{Model} 
& $C$ & $S$ & $\Delta$ 
& FID$_{X\!\to\!Y}$ & FID$_{Y\!\to\!X}$ & Acc. \\
\midrule

\textbf{Base ($\mathcal{L}_{\text{lat}} \!+\! \mathcal{L}_{\text{img}}$)}
  & 0.74 $\pm$ 0.010 & 0.73 $\pm$ 0.02 & 0.51
  & 39.78 & 55.75 & 0.535 \\

\textbf{Base + $\mathcal{L}_{D}$} 
  & 0.64 $\pm$ 0.02 & 0.92 $\pm$ 0.01 & 0.22
  & 37.03 & 53.21 & 0.638 \\

\textbf{Base + $\mathcal{L}_{D} \!+\! \mathcal{L}_{\text{DiscMI}}$} 
  & 0.68 $\pm$ 0.01 & 0.86 $\pm$ 0.01 & 0.32
  & 38.78 & 55.04 & 0.593 \\

\textbf{Base + $\mathcal{L}_{D} \!+\! \mathcal{L}_{R}$} 
  & \textbf{0.58 $\pm$ 0.08} & \textbf{0.95 $\pm$ 0.01} & \textbf{0.13}
  & 37.10 & 45.50 & \textbf{0.939} \\
  
\textbf{Base + $\mathcal{L}_{D} \!+\! \mathcal{L}_{R} \!+\! \mathcal{L}_{\text{refine}}$} 
  & -- & -- & --
  & \textbf{25.17} & \textbf{35.56} & \uline{0.913} \\

\bottomrule
\end{tabular}%
}
\end{minipage}
\end{table}

\begin{table*}[t]
\centering
\caption{\textbf{CF explanation comparison on BraTS.} Best in bold, second-best underlined. Metrics for CF edits (X$\rightarrow$Y: add tumors; Y$\rightarrow$X: remove tumors) and runtime.}
\label{tab:cf_eval}
\footnotesize
\resizebox{0.85\textwidth}{!}{%
\begin{tabular}{l|cccccc|cccccc|c}
\toprule
& \multicolumn{6}{c|}{\textbf{CF: X$\rightarrow$Y (add tumors)}} & \multicolumn{6}{c|}{\textbf{CF: Y$\rightarrow$X (remove tumors)}} & \\
\cmidrule(lr){2-7}\cmidrule(lr){8-13}
\textbf{Method}
& FID & sFID & L1 & FR & MAD & BKL
& FID & sFID & L1 & FR & MAD & BKL
& Time (s/img) \\
\midrule
ACE & 38.553 & 51.428 & \textbf{0.003} & \textbf{0.977} & \uline{0.942} & \textbf{0.043}
    & \uline{39.436} & \uline{52.841} & \textbf{0.005} & \uline{0.895} & \uline{0.855} & \uline{0.120}
    & 21.37 $\pm$ 0.34  \\
DiME & 38.378 & 50.421 & 0.026 & 0.947 & 0.934 & 0.054
     & 48.752 & 60.722 & 0.025 & 0.741 & 0.723 & 0.228
     & 294.12 $\pm$ 47.44 \\
FastDiME & \uline{32.431} & \uline{44.735} & 0.017 & 0.905 & 0.873 & 0.108
        & 51.391 & 64.007 & 0.016 & 0.696 & 0.658 & 0.285
        & 16.07 $\pm$ 0.64 \\
FastDiME-2+ & 32.865 & 45.283 & 0.017 & 0.912 & 0.876 & 0.106
           & 52.191 & 64.753 & 0.016 & 0.707 & 0.668 & 0.277
           & 30.69 $\pm$ 0.91 \\
TIME & 76.704 & 91.064 & \uline{0.012} & 0.389 & 0.293 & 0.475
     & 88.706 & 100.074 & \uline{0.012} & 0.341 & 0.247 & 0.413
     & 17.59 $\pm$ 0.40 \\
SDXL-LoRA & 51.258 & 60.179 & 0.027 & 0.731 & 0.727 & 0.230
          & 59.006 & 69.185 & 0.028 & 0.824 & 0.817 & 0.164
          & 4.95 $\pm$ 0.67 \\
\midrule
Ours & \textbf{27.533} & \textbf{38.441} & 0.021 & \uline{0.953} & \textbf{0.946} & \uline{0.047}
     & \textbf{37.492} & \textbf{47.527} & 0.022 & \textbf{0.914} & \textbf{0.906} & \textbf{0.074}
     & \textbf{0.25 $\pm$ 0.005} \\
\bottomrule
\end{tabular}%
}
\end{table*}

\begin{figure}[t]
    \centering
    \includegraphics[width=0.95\linewidth]{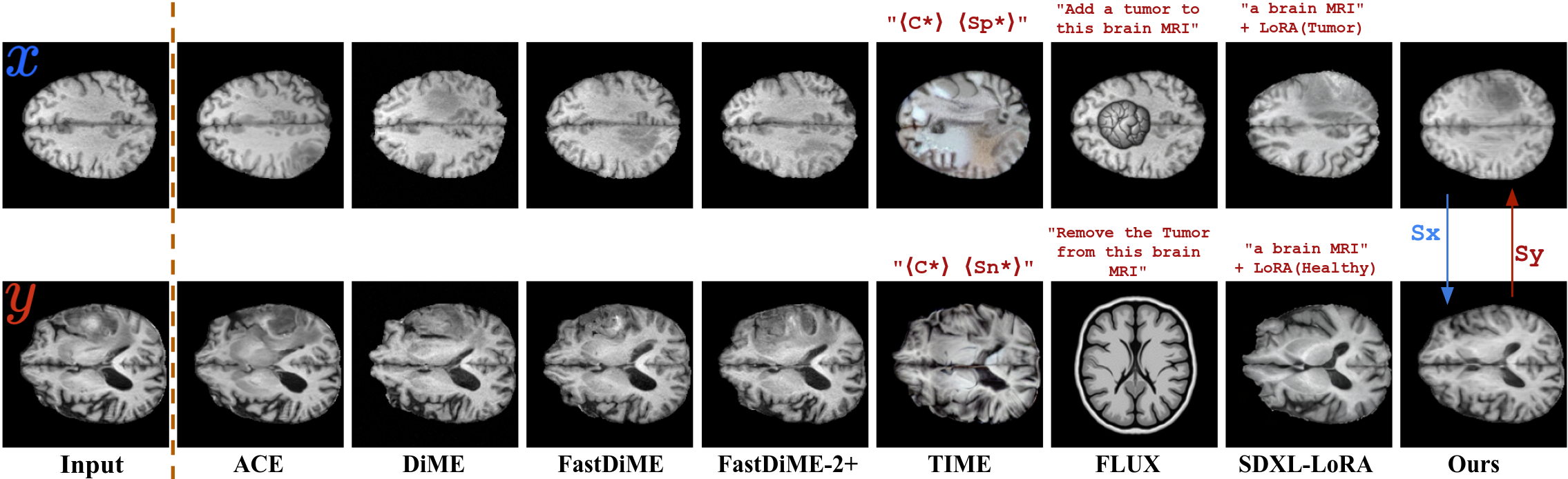}
    \caption{\textbf{Comparisons of CF generation.} Cols. 2--5 show CF outputs from SOTA diffusion-based methods. Cols. 6--8 show CF outputs from T2I diffusion models.}
    \label{fig:cf_comp}
\end{figure}

\noindent \textbf{Counterfactual Explanations Based on CA.}
Table~\ref{tab:cf_eval} reports quantitative comparisons on BraTS Healthy ($X$) and Tumored ($Y$) for two counterfactual directions: adding tumors ($X\!\rightarrow\!Y$) and removing tumors ($Y\!\rightarrow\!X$). 
Overall, our CA-based swapping achieves the best image fidelity in both directions, as indicated by the lowest FID/sFID scores, and is better or on par with diffusion-based CF baselines (ACE, DiME, and FastDiME) in terms of counterfactual effectiveness (i.e., FR/MAD/BKL). 
In Fig.~\ref{fig:cf_comp}, we observe 
that diffusion-based methods struggle to completely remove large tumors from $Y$ images, and the added tumors in $X\!\rightarrow\!Y$ have often limited size and/or are misplaced. 
In addition, text-conditioned editing (e.g., FLUX) or learning per-image embeddings (e.g., TIME) can drift away from the real image manifold, while fine-tuning large diffusion models with LoRA (e.g., SDXL-LoRA) may alter common content beyond the desired tumor-related changes. 
By contrast, our approach explicitly swaps salient generative factors between $X$ and $Y$, improving image fidelity and offering better controllability for tumor-specific counterfactual manipulation.

\begin{figure}[t]
    \centering
    \includegraphics[width=0.9\linewidth]{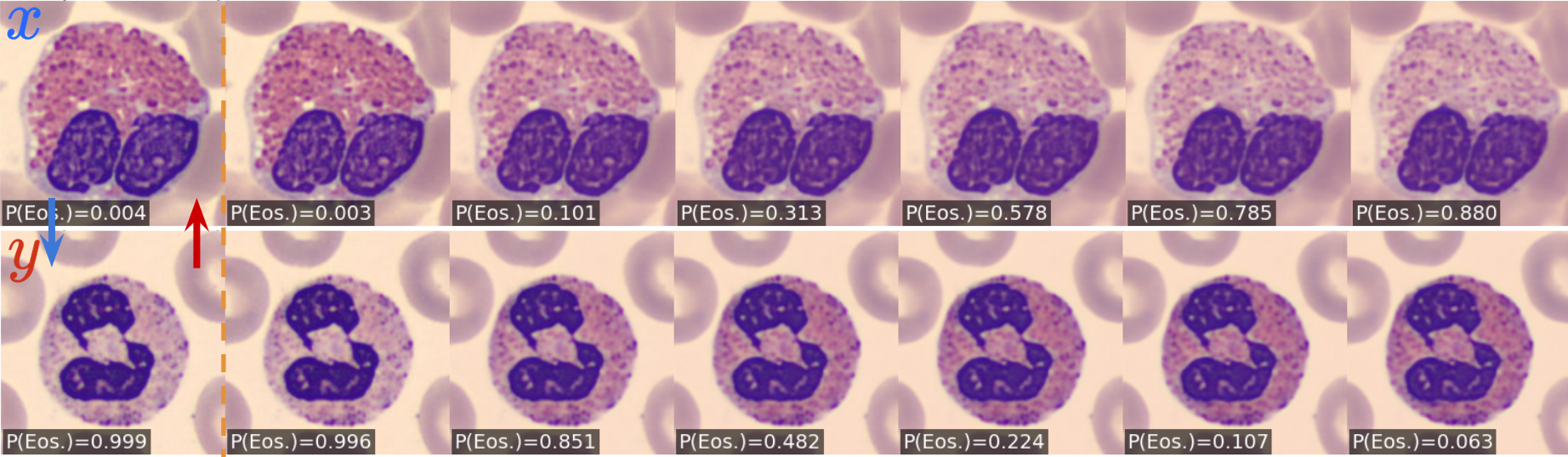}
    \caption{\textbf{Interpolation along salient factors.}
    Gradual interpolation from $s_x$ to $s_y$ (and viceversa) using an interpolation weight $\alpha$ ($0\!\rightarrow\!1$) during generation. The classifier prediction changes consistently along the trajectory, illustrating gradual insertion/removal of class-specific evidence while preserving common content.}
    \label{fig:interp}
\end{figure}

\noindent Thanks to the well structured latent space of StyleGAN2, we can also interpolate between salient factors.
Given a pair of samples $(x,y)$ from two domains/classes, we can generate a continuous counterfactual trajectory by scaling the salient change with a scalar $\alpha\in[0,1]$, e.g., $G(\alpha\widehat{f}_{x \rightarrow y})$. This is shown in Fig.~\ref{fig:interp}, where the predicted class probability changes smoothly along the trajectory, suggesting that the generated counterfactuals form a coherent explanation path.

\section{Conclusion}
We propose a classifier-free, CA-based framework for VCEs that yields strong common–salient disentanglement and high-quality CF generations. Interesting perspectives might be adapting the proposed method to recent VLM, as in \cite{kazimi_explaining_2025}, and studying its identifiability. Similarly to all recent VCE \cite{jeanneret2022diffusion,Jeanneret_2023_CVPR,weng2024fast} and CA \cite{carton2024double, louiset24a} works, we note that also our method also, without strong assumptions, is not identifiable \cite{hyvarinen19a,locatello19a} and does not enable causal discovery. This is generally impossible from purely observational single-modal data. An interesting direction, in line with causal representation learning theory \cite{scholkopf_statistical_2022}, would be using additional prior or longitudinal data, to further constrain the learning process.

\begin{credits}
\subsubsection{\ackname} 
This work was performed using HPC resources from
GENCI–IDRIS (A0160615058) and it was supported by the
EUR BERTIP (ANR-18-EURE-0002).


\end{credits}
\bibliographystyle{splncs04}
\bibliography{mybibliography}

\clearpage

\begin{center}
{\LARGE\bfseries Supplementary Material\par}
\end{center}

\makeatother


\section{Architecture Details}
\label{sec:arch_details}
\subsection{CS Separator}
\label{sec:cs_separator_arch}

As described in Section~\ref{sec:methods}, the CS separator $H_{\mathrm{cs},\phi}$ contains one common branch and one or two salient branches, depending on the contrastive analysis setting. In the \textit{background--target} (BT) setting, it consists of the common branch $C_{\phi_c}$ and the salient branch $S_{y,\phi_{s_y}}$. In the \textit{multiple-salient} (MS) setting, an additional salient branch $S_{x,\phi_{s_x}}$ is introduced to capture features specific to dataset
$X$. All branches share the same architecture but have independent parameters.

Following He et al.~\cite{he2025learningcommonsalientgenerative}, we implement each branch as an MLP adapted from StyleGAN2's $\mathcal{Z}$-to-$\mathcal{W}$ mapping network~\cite{Karras2019stylegan2}. Each branch comprises four linear layers with equalized learning-rate scaling, each followed by a LeakyReLU activation with a negative slope of $0.2$. Unlike the original StyleGAN2 $\mathcal{Z}$-to-$\mathcal{W}$ mapping network, in which each linear layer applies a single $512\times512$ weight matrix to a 512-dimensional latent vector, each linear layer in our separator is parameterized by a weight tensor $\mathbf{A}^{(k)}\in\mathbb{R}^{L\times512\times512}$ to process the $\mathcal{W}^{+}$ code $w=[w_1,\ldots,w_L]^\top\in\mathbb{R}^{L\times512}$. Here, $k=1,\ldots,4$ indexes the linear layer, and $L$ denotes the number of style vectors in the $\mathcal{W}^{+}$ code. For a generator with a native output resolution of $R\times R$, $L$ is given by $L=2\log_2(R)-2$. In our experiments, all StyleGAN2 generators are pretrained at a native resolution of $256\times256$, yielding $L=14$. This design assigns a position-specific weight matrix to each style position. Since each layer preserves the dimensionality of the latent code, it maps from $\mathbb{R}^{B\times L\times512}$ to $\mathbb{R}^{B\times L\times512}$, where $B$ denotes the batch size.

\subsection{Regularization Networks $D$ and $R$}
\label{sec:regularization_networks}

As described in Section~\ref{sec:methods}, we employ a domain
discriminator $D$ and a dependency regressor $R$ to regularize the
common and salient representations. Given a common representation
$c\in\mathbb{R}^{L\times512}$, the domain discriminator first flattens
the complete representation into a single vector. Since our StyleGAN2
generator operates at a native resolution of $256\times256$, we have
$L=14$, resulting in a 7168-dimensional input vector. The discriminator is implemented as a binary linear classifier with a single fully connected layer mapping the 7168-dimensional input to one output logit. No hidden
layer or output activation is used, and the network directly produces a
single domain-classification logit.

The dependency regressor $R$ predicts the salient representation from
the corresponding common representation. Given a batch of common representations
$c\in\mathbb{R}^{B\times L\times512}$, the batch and style dimensions
are merged to form a matrix in $\mathbb{R}^{(BL)\times512}$. The
regressor consists of two fully connected layers that map this matrix
from $(BL)\times512$ to $(BL)\times512$, with a ReLU activation between
the two layers and no activation after the output layer. The output is then reshaped back to obtain the predicted salient factor
$\hat{s}\in\mathbb{R}^{B\times L\times512}$.

\subsection{Other Architectures and Implementations}

For the remaining network components, we use the pSp architecture~\cite{richardson2021encoding} for $E$ and the StyleGAN2 architecture~\cite{Karras2019stylegan2} for $G$. The F-space refinement stage builds upon SFE~\cite{bobkov2024devil}, retaining the original architectures of its F-space encoder $E_f$ and trainable adapter $F_{\mathrm{Adpt}}$. For the original StyleGAN2 $\mathcal{Z}$-to-$\mathcal{W}$ mapping network on which the separator architecture is based, we refer to the implementation provided at \url{https://nn.labml.ai/gan/stylegan/index.html}. For the StyleGAN2 generator $G$, we use the PyTorch implementation by rosinality, available at \url{https://github.com/rosinality/stylegan2-pytorch}. The same pretrained $G$ is used throughout both training stages and evaluation. To compute $\mathcal{L}_{\mathrm{adv}}$, we use the discriminator pretrained jointly with $G$.

For image-level evaluation, we adopt the U-Net-based classifier used in~\cite{weng2024fast}. The classifier is trained on real images from the corresponding training set and subsequently applied to the edited images to assess whether they exhibit the intended target attributes. For latent-space evaluation, we quantify the separation between the learned common and salient factors by training a logistic regression classifier to predict the dataset label from each factor. The classifier is implemented in scikit-learn. Each factor is flattened and normalized before being provided to the classifier.

\section{Additional Results}
\begin{figure}[t]
    \centering
    \includegraphics[width=0.8\linewidth]{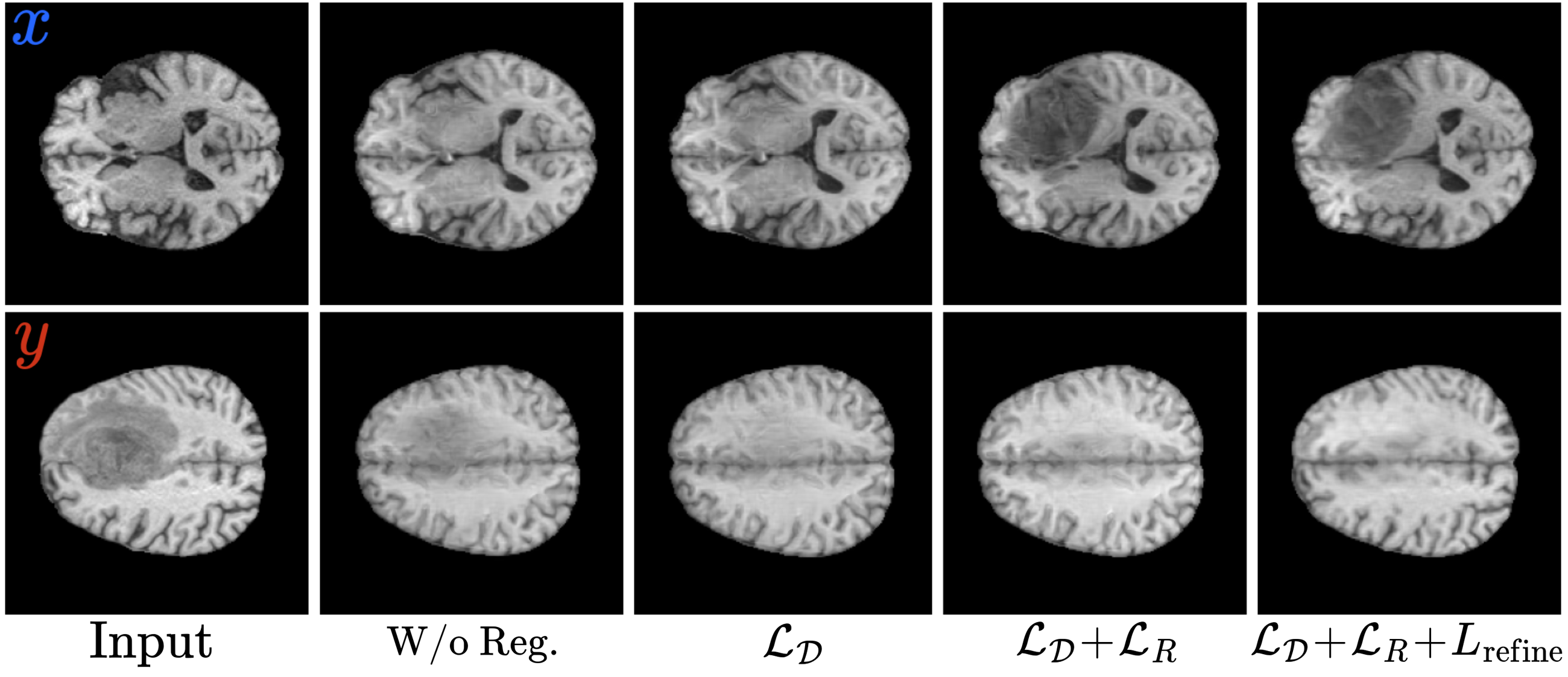}
\caption{\textbf{Qualitative ablation study of the training components on BraTS}. The first column shows inputs $x\in X$ and $y\in Y$ from the healthy and tumor datasets, respectively. Columns 2--5 show the corresponding salient-factor swapping results under different training configurations.}
    \label{fig:ablation_comp} 
\end{figure}

\begin{figure}[t]
    \centering
    \includegraphics[width=0.8\linewidth]{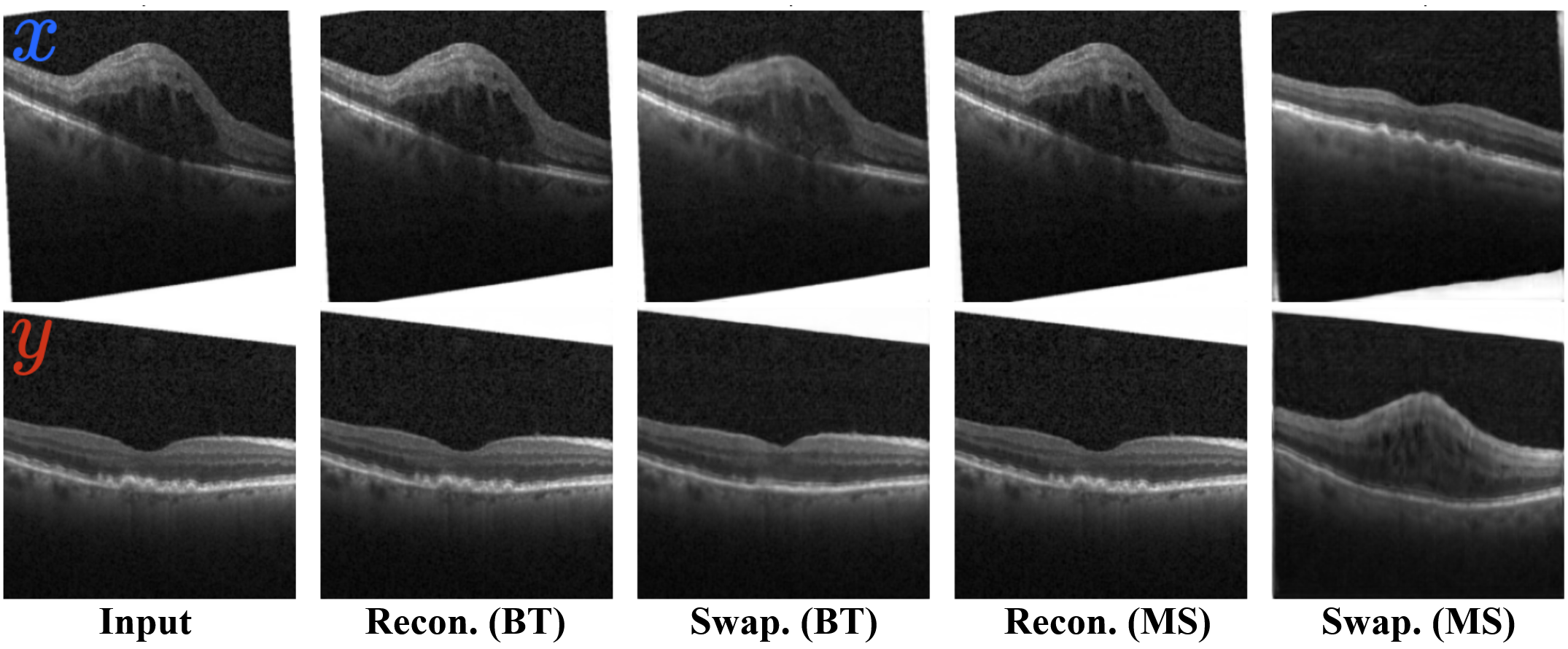}
\caption{\textbf{Comparison of the BT and MS assumptions on OCTMNIST}. The input images $x$ and $y$ are sampled from the DME and drusen datasets, respectively. The remaining columns show the reconstruction and salient-factor swapping results under the BT and MS assumptions.}
\label{fig:ablation_ms}
\end{figure}

Fig.~\ref{fig:ablation_comp} illustrates the effects of different
training components on salient-factor swapping. With only the base
reconstruction objectives (W/o Reg.), the swaps produce limited class-specific
changes. Adding $\mathcal{L}_{D}$ improves tumor removal in the
$y\rightarrow x$ direction but not tumor addition in the
$x\rightarrow y$ direction. Adding $\mathcal{L}_{R}$ improves
common--salient separation, enabling clearer bidirectional edits while
better preserving the source anatomy. Finally, F-space refinement
enhances local details and image fidelity without changing the semantic
effect of the edit. These results are consistent with the quantitative
ablation in Table~3.
Fig.~\ref{fig:ablation_ms} compares the BT and MS assumptions on
OCTMNIST. While both variants reconstruct the inputs well, the BT
variant produces limited swapping effects because it cannot separately
model the class-specific patterns of DME and drusen. The proposed MS variant,
with two salient branches, enables more meaningful bidirectional swaps
while preserving the common retinal structure.

\begin{figure}[t]
    \centering
    \includegraphics[width=0.8\linewidth]{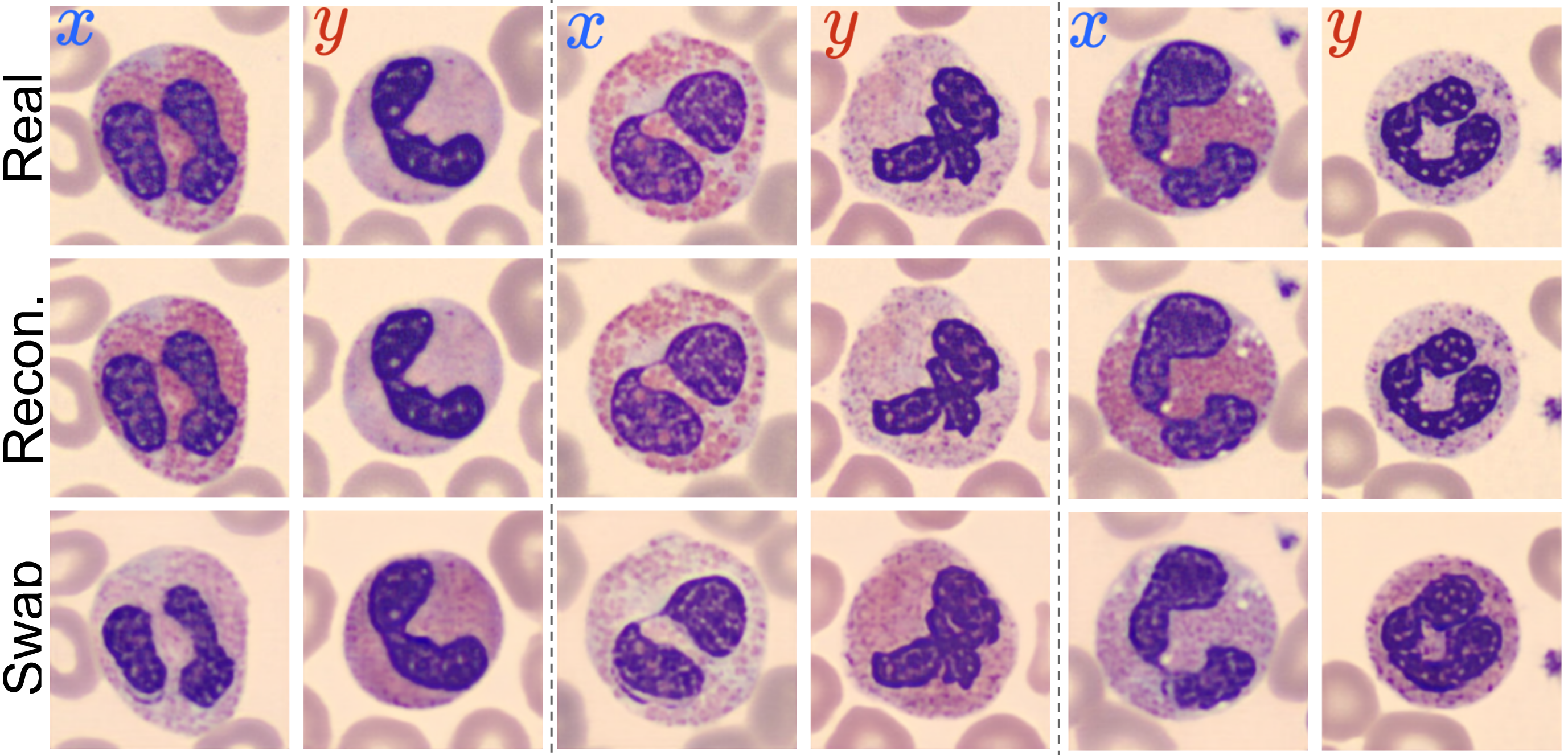}
\caption{\textbf{Additional examples on BloodMNIST under the BT
assumption}. The input images $x$ and $y$ are sampled from the
eosinophil and neutrophil datasets, respectively.}
\label{fig:blood_additional}
\end{figure}

\begin{figure}[t]
    \centering
    \includegraphics[width=0.8\linewidth]{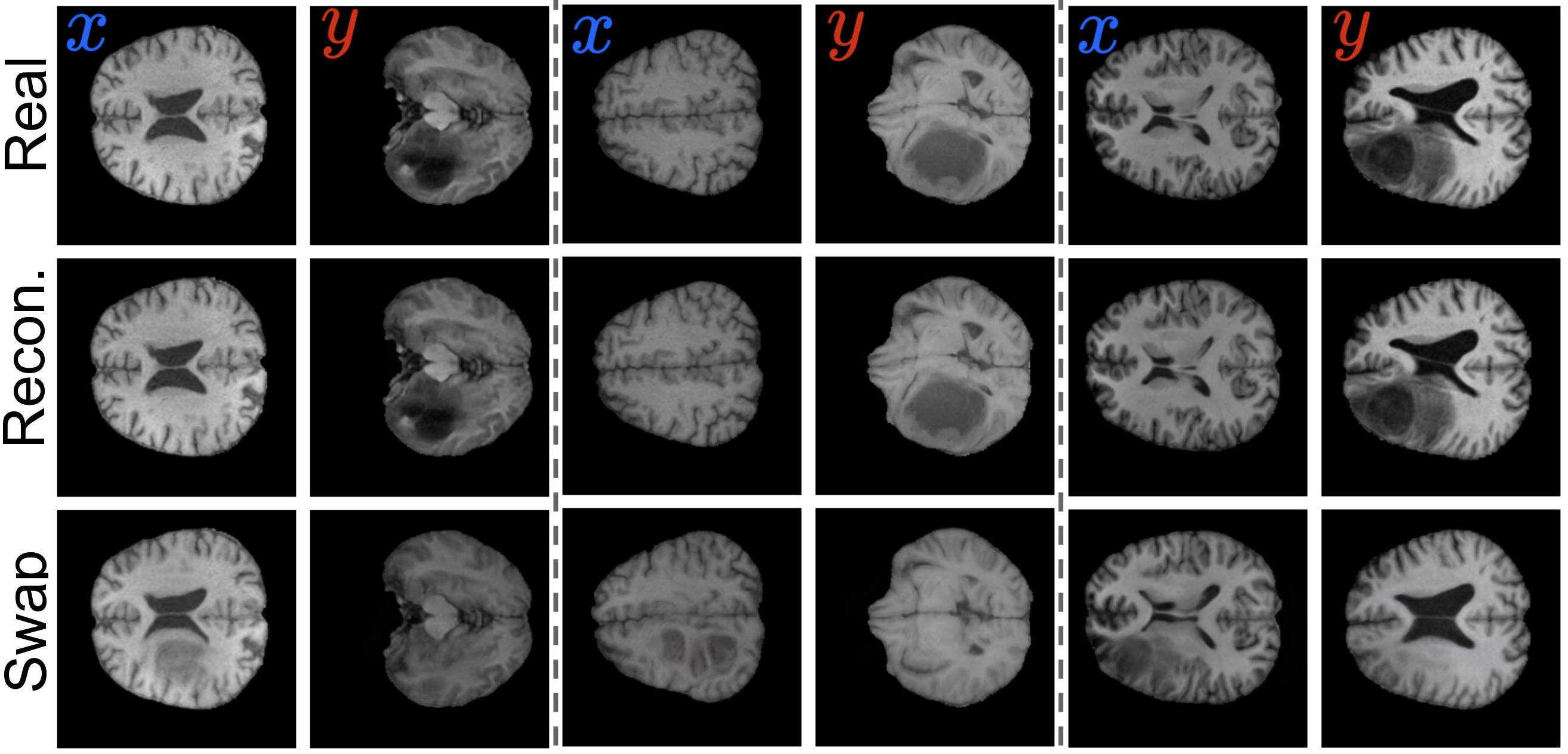}
\caption{\textbf{Additional examples on BraTS under the BT
assumption}. The input images $x$ and $y$ are sampled from the healthy
and tumor MRI scans, respectively.}
\label{fig:mri_additional}
\end{figure}

\begin{figure}[t]
    \centering
    \includegraphics[width=0.8\linewidth]{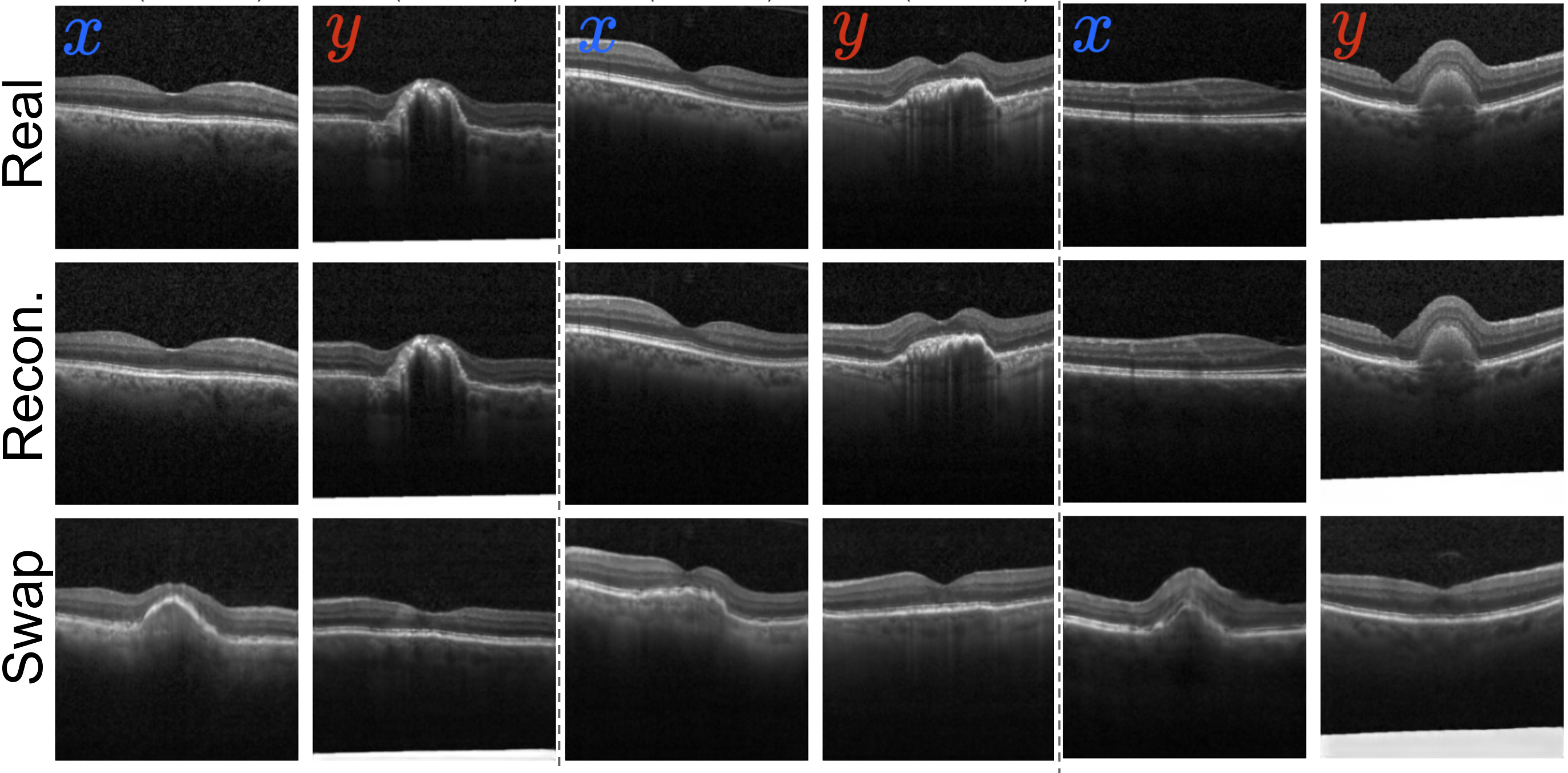}
\caption{\textbf{Additional examples on OCTMNIST under the BT
assumption}. The input images $x$ and $y$ are sampled from the normal
and choroidal neovascularization (CNV) datasets, respectively.}
\label{fig:oct_bt_additional}
\end{figure}

\begin{figure}[t]
    \centering
    \includegraphics[width=0.8\linewidth]{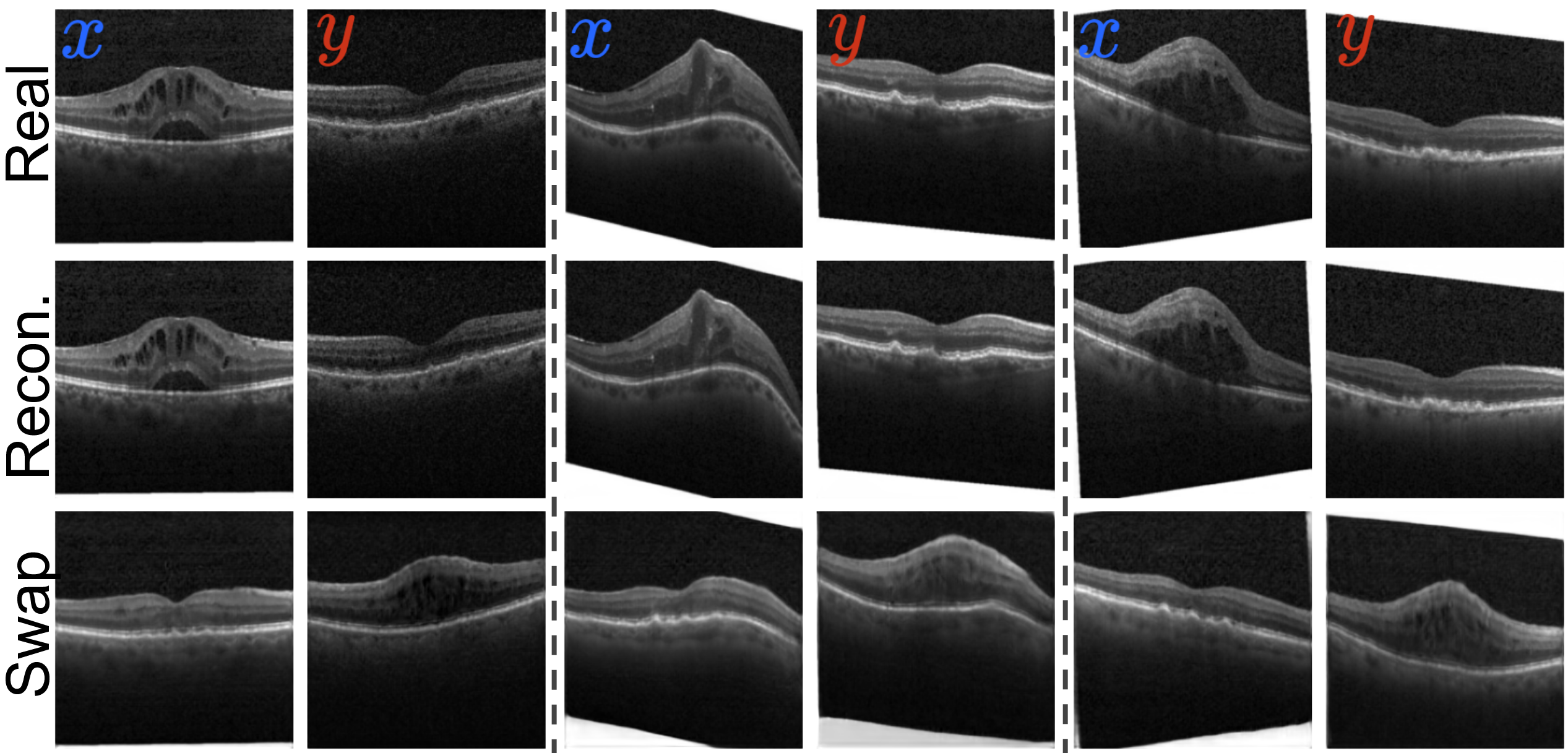}
\caption{\textbf{Additional examples on OCTMNIST under the MS assumption}. The input images labeled $x$ and $y$ are sampled from the DME and drusen datasets, respectively.}
\label{fig:oct_ms_additional}
\end{figure}

\begin{figure}[t]
    \centering
    \includegraphics[width=1.0\linewidth]{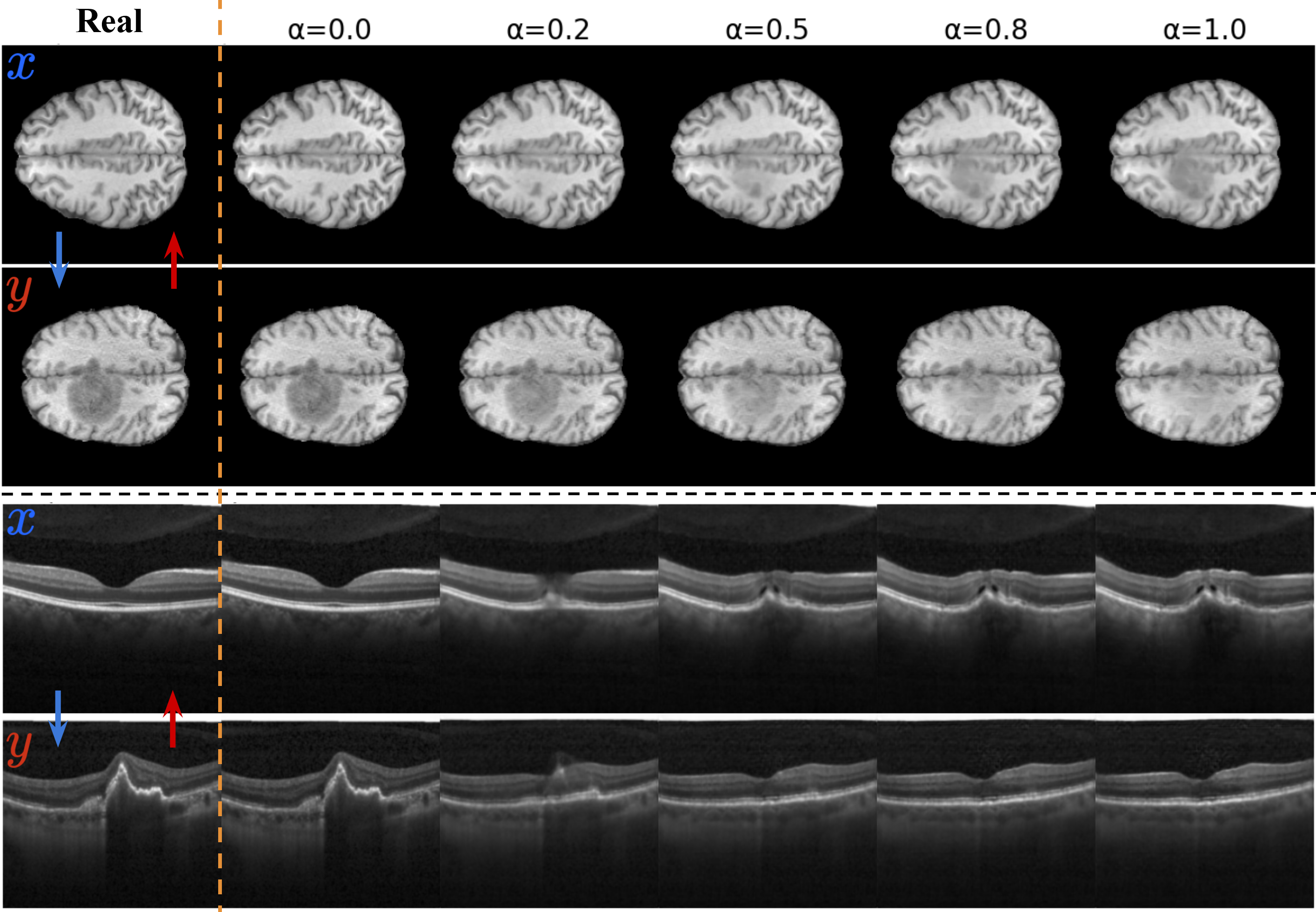}
\caption{\textbf{Additional interpolation results on BraTS and OCTMNIST}. 
The upper two rows correspond to a healthy image $x$ and a tumor image $y$ from BraTS, and the lower two rows to a normal image $x$ and a CNV image $y$ from OCTMNIST. The first row of each pair interpolates from $s_x$ to $s_y$ while preserving $c_x$, and the second interpolates from $s_y$ to $s_x$ while preserving $c_y$.}
\label{fig:interpolation_additional}
\end{figure}

Figs.~\ref{fig:blood_additional}--\ref{fig:oct_ms_additional} show
additional reconstruction and salient-factor swapping results. Each
pair of columns, separated by dashed lines, corresponds to an input pair
$(x,y)$. From top to bottom, the rows show the real images, their
reconstructions, and the swapping results. The swapped images are
obtained by combining the common representation of each source image
with the salient representation of the other image. Figs.~\ref{fig:blood_additional}--\ref{fig:oct_bt_additional}
use the BT assumption, whereas Fig.~\ref{fig:oct_ms_additional} uses
the MS assumption. Under the BT
assumption, the salient factors of $x$ is absent, whereas the MS
assumption models separate salient factors for both $x$ and
$y$. Figure~\ref{fig:interpolation_additional} provides additional interpolation examples on BraTS and OCTMNIST.

\end{document}